\documentclass[10pt,twocolumn,letterpaper]{article}

\usepackage{cvpr}              

\usepackage{graphicx}
\usepackage{amsmath}
\usepackage{amssymb}
\usepackage{booktabs}

\usepackage[pagebackref,breaklinks,colorlinks]{hyperref}
\usepackage{caption}
\newcommand{\gray}[1]{\textcolor{gray}{{#1}}}
\newcommand{\green}[1]{\color[HTML]{3166FF}{{#1}}}
\newcommand{\white}[1]{\color[HTML]{FFFFFF}{{#1}}}

\newcommand{\figref}[1]{Fig.~\ref{#1}}

\newcommand{\tabref}[1]{Table~\ref{#1}}

\newcommand{\secref}[1]{Sec.~\ref{#1}}
\renewcommand{\paragraph}[1]{\vspace{1mm}\noindent\textbf{#1}}
\usepackage{multirow}
\usepackage{tabu}

\usepackage[capitalize]{cleveref}
\crefname{section}{Sec.}{Secs.}
\Crefname{section}{Section}{Sections}
\Crefname{table}{Table}{Tables}
\crefname{table}{Tab.}{Tabs.}

\usepackage{multirow}
\usepackage[table,xcdraw]{xcolor}

\makeatletter
\@namedef{ver@everyshi.sty}{}
\makeatother
\usepackage{tikz}

\usepackage{subcaption}

\usepackage{tikz}
\usetikzlibrary{arrows,intersections}
\tikzset{font=\scriptsize}
\usepackage{pgfplots}
\pgfplotsset{compat=1.11}
\usepgfplotslibrary{fillbetween}
\usetikzlibrary{backgrounds}

\usepackage{xspace}
\usepackage{tabu}
\newlength\savewidth\newcommand\shline{\noalign{\global\savewidth\arrayrulewidth
  \global\arrayrulewidth 1pt}\hline\noalign{\global\arrayrulewidth\savewidth}}
\newcommand{\tablestyle}[2]{\setlength{\tabcolsep}{#1}\renewcommand{\arraystretch}{#2}\centering\footnotesize}

\newcommand{\ours}{RO-ViT\xspace}

\makeatletter
\newcommand{\thickhline}{%
    \noalign {\ifnum 0=`}\fi \hrule height 1pt
    \futurelet \reserved@a \@xhline
}
\makeatother

\usepackage{bm}

\renewcommand{\Sigma}{\mathfrak{S}}

\def\1{\bm{1}}

\DeclareMathAlphabet{\mathsfit}{\encodingdefault}{\sfdefault}{m}{sl}
\SetMathAlphabet{\mathsfit}{bold}{\encodingdefault}{\sfdefault}{bx}{n}

\def\cvprPaperID{9093} 
\def\confName{CVPR}
\def\confYear{2023}

\begin{document}
\title{Region-Aware Pretraining for Open-Vocabulary Object Detection with \\ Vision Transformers}
\author{
Dahun Kim \quad\quad\quad Anelia Angelova \quad\quad\quad Weicheng Kuo\\
Google Research, Brain Team\\
}

\maketitle

\begin{abstract}
We present Region-aware Open-vocabulary Vision Transformers (\ours)
– a contrastive image-text pretraining recipe to bridge the gap between image-level pretraining and open-vocabulary object detection. 
At the pretraining phase, we propose to randomly crop and resize regions of positional embeddings instead of using the whole image positional embeddings. This better matches the use of positional embeddings at region-level in the detection finetuning phase. In addition, we replace the common softmax cross entropy loss in contrastive learning with focal loss to better learn the informative yet difficult examples. Finally, we leverage recent advances in novel object proposals to improve open-vocabulary detection finetuning. We evaluate our full model on the LVIS and COCO open-vocabulary detection benchmarks and zero-shot transfer. \ours achieves a state-of-the-art 34.1 $AP_r$ on LVIS, surpassing the best existing approach by +7.8 points in addition to competitive zero-shot transfer detection. Surprisingly, \ours improves the image-level representation as well and achieves the state of the art on 9 out of 12 metrics on COCO and Flickr image-text retrieval benchmarks, outperforming competitive approaches with larger models.\footnote{project page: \href{https://github.com/mcahny/rovit}{https://github.com/mcahny/rovit}}
\end{abstract}
\vspace{-3mm}
\section{Introduction}
\label{sec:intro}

\begin{figure}[t]
\begin{center}
\includegraphics[width=0.9\linewidth]{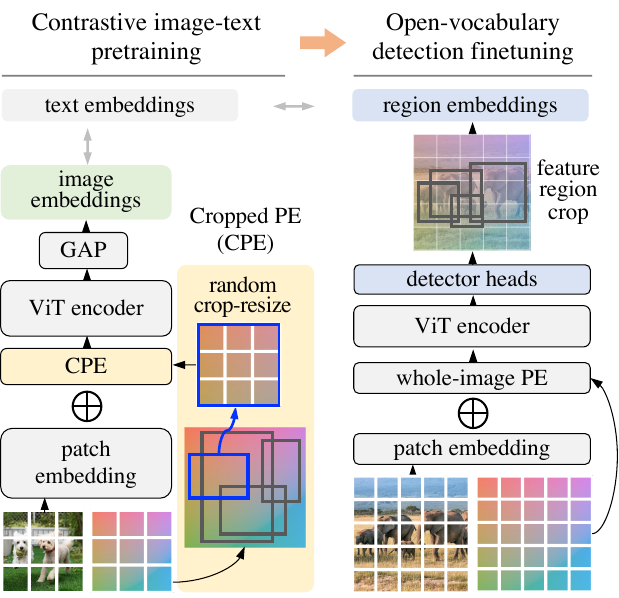}
\vspace{-1mm}
\caption{Existing vision-language models are designed for image-level tasks, \eg, classification and retrieval. In this paper, we aim to enhance the image-text pretraining for region-level downstream task: open-vocabulary object detection. At pretraining, we propose to randomly crop and resize regions of positional embeddings (PE) instead of using the whole image PE. This better matches the use of PE at region-level in the detection finetuning and results in better performance in detection and surprisingly also retrieval task.
}
\label{fig:teaser}
\vspace{-6mm}
\end{center}
\end{figure}

The ability to detect objects in the visual world is a hallmark of computer vision and machine intelligence. It is key to enable many applications, e.g. autonomous agents adapting to new environments with many novel objects, a shopping system handling fine-grained user queries such as ``fedora'', ``bonnet'' outside of the training vocabulary. However, modern object detectors typically rely on manual annotations of regions and class labels, which limits their vocabulary size to an order of $10^3$ and it is prohibitively expensive to scale up further. This is orders of magnitude smaller than the objects we encounter in the visual world. 

Recently, the open-vocabulary detection task (OVD) has been proposed to overcome such limitation by leveraging abundant image-text pairs for training and ingesting the text queries provided by users at test time~\cite{Zareian_2021_CVPR}. By treating the categories as text embeddings rather than discrete ids, open-vocabulary detectors can flexibly predict a wide range of objects unseen during the training time. Most existing works leverage image-text pre-training which contains rich semantic knowledge of open-vocabulary concepts. Knowledge distillation~\cite{gu2022openvocabulary,du2022learning}, weak supervision~\cite{zhou2022detecting}, self-training~\cite{zhong2021regionclip,rasheed2022bridging,zhao2022exploiting}, and frozen model~\cite{kuo2022f} have been proposed, and CNN backbones are most commonly used. With the growing popularity of vision transformers in image understanding~\cite{dosovitskiy2020image,li2022exploring,zhang2022segvit}, multimodal~\cite{arnab2021vivit,flamingo,wang2021simvlm}, and self-supervised tasks~\cite{He_2022_CVPR,bao2021beit,Caron_2021_ICCV}, it is important to understand how to build capable open-vocabulary detectors with vision transformers~\cite{dosovitskiy2020image,minderer2022simple}.

To our best knowledge, all existing works assume pretrained Vision-Language Models (VLM) are given, and develop adaptation or finetuning recipes to bridge the gap between image-level pretraining and object-level finetuning~\cite{gu2022openvocabulary,du2022learning,zhong2021regionclip,zhao2022exploiting,rasheed2022bridging}. Since the VLMs are designed for image-level tasks \eg classification, retrieval, the notion of objects/regions are not adequately utilized in the pretraining process. We believe it would be beneficial for open-vocabulary detection if we bake in locality information in the image-text pretraining.

We present \ours, a simple recipe to pretrain vision transformers in a region-aware manner for open-vocabulary object detection. Standard pretraining typically uses full-image positional embeddings, which does not generalize well to detection tasks. Thus, we propose a novel positional embedding scheme called ``Cropped Positional Embedding" which better matches the use of region crops in detection finetuning (see \figref{fig:teaser}). In addition, we found it beneficial to replace the softmax cross entropy loss with focal loss in contrastive learning, which affords us more control to learn from harder and more informative examples. Finally, we leverage recent advances in novel object proposals~\cite{kim2022learning} to improve open-vocabulary detection finetuning. The motivation is that existing approaches often miss novel objects in the object proposal stage because the proposals tend to overfit to the foreground categories. 

We evaluate the approach on the standard LVIS and COCO open-vocabulary benchmarks. On LVIS, our best model achieves a state-of-the-art 34.1 AP$_r$ at the system level, surpassing the best existing approach by +7.8 AP$_r$. Compared to the best existing ViT-based approach, ours outperforms by a healthy margin of +9.5 AP$_r$. Our LVIS-trained model outperforms existing baselines on transfer detection to Objects365 without re-training. Although not explicitly optimized for retrieval, \ours surprisingly achieves the state-of-the-art performance on 9 out of 12 metrics in image-text retrieval benchmark and outperforms strong baselines with standard positional embeddings, cross entropy loss, and larger model capacity. We conduct ablations to confirm the benefits of each proposed component. Visualization of the learnt positional embeddings shows that our approach (CPE) leads to more symmetrical and diverse patterns than the baseline. In summary: \vspace{-1mm}
\begin{itemize}
    \item We propose \ours to bridge the positional embeddings in image-text pretraining to open-vocabulary detection finetuning.\vspace{-2mm}
    \item We show that image-text pretraining with focal loss is more effective than existing softmax CE loss.\vspace{-2mm}
    \item We improve the open-vocabulary detection finetuning recipe with novel object proposals.\vspace{-2mm}
    \item \ours achieves the state of the art on the LVIS open-vocabulary detection benchmark, and 9 out 12 metrics on COCO and Flickr image-text retrieval benchmarks.
\end{itemize}

We hope these findings will facilitate the research community to further explore open-vocabulary detection from the perspective of image-text pretraining with potential benefits for both region-level and image-level tasks. 

\section{Related Works}
\label{sec:related}

\paragraph{Learning open-vocabulary and zero-shot recognition.}\quad
Building general representation for open-vocabulary and zero-shot recognition is a fundamental machine learning task. DeViSE~\cite{frome2013devise} and ConSE~\cite{norouzi2013zero} are pioneering works to learn a shared image-text embedding space for zero-shot recognition by convolutional networks. As image and language often co-occur on the internet, the community has explored learning such representation from various data sources including image tags~\cite{chen2015webly,divvala2014learning,joulin2016learning}, captions/text descriptions~\cite{desai2021virtex,he2017fine,sariyildiz2020learning,wang2009learning,zhong2021learning}, alt-texts~\cite{align,basic,zhai2021lit}, and image search queries~\cite{radford2021clip}. Recently, contrastive learning has become a popular paradigm to learn from large image-text corpus due to its simplicity and scalability, where increasing model size and data yield consistent improvement on open-vocabulary classification benchmarks. The aforementioned works focus on image-level recognition, whereas we study how to best tailor them for region-level understanding.

\paragraph{Self-supervised representation learning for localization.}\quad
Due to the challenge to scale up labeling for localization tasks, many efforts have been made to learn locality-sensitive representation in a self-supervised manner. Existing approaches roughly fall into the contrastive or generative paradigms. Contrastive approaches typically involve region or point-level contrastive learning using sliding windows~\cite{Xiao_2021_ICCV}, object proposals~\cite{SoCo,Henaff_2021_ICCV}, or point samples~\cite{Bai_2022_CVPR}. Most existing contrastive methods are CNN-based, while ViT backbones have been popular recently with generative approaches~\cite{He_2022_CVPR,bao2021beit} and has shown promise with contrastive approaches~\cite{Caron_2021_ICCV}. Masked image modeling is commonly used to learn the ViT features for localization tasks. Even though these self-supervised methods are well-suited for standard localization tasks, they lack the image-text correspondence necessary for open-vocabulary recognition. Thus, we study contrastive image-text representation learning for open-vocabulary detection with ViTs.

\paragraph{Open-vocabulary object detection and segmentation.}\quad
Zero-shot detection was proposed to scale up detection models beyond their limited training categories. Popular approaches accomplish this by learning the alignment between region visual representation and category word embeddings~\cite{bansal2018zero, rahman2020improved, demirel2018zero, zheng2020background} or hallucinating visual features with a generative model~\cite{hayat2020synthesizing,zhu2020don}. Without any visual knowledge of the novel categories, zero-shot detectors tend to struggle with low performance. This motivates the open-vocabulary detection benchmark~\cite{Zareian_2021_CVPR} to bridge the gap between zero-shot and fully-supervised detection.

With the rise of image-text pretraining, many works have explored adapting these pretrained models to open-vocabulary detection and segmentation ~\cite{gu2022openvocabulary,zhong2021regionclip,ghiasi2022scaling,li2022language,zhou2022maskclip}. For example, ViLD~\cite{gu2022openvocabulary} proposes to distill the image-text knowledge into the detector. DetPro~\cite{du2022learning} improves ViLD by incorporating the idea of prompt optimization~\cite{zhou2022cocoop}. Moreover, region-text self-training has been shown effective on image caption data~\cite{zhong2021regionclip}, classification data~\cite{rasheed2022bridging}, or even unlabeled data~\cite{zhao2022exploiting}. Weak supervision~\cite{zhou2022detecting} and phrase grounding~\cite{li2021grounded} approaches have also been proposed. In terms of architecture, CNNs are most commonly used but vision transformers have recently been adopted as well~\cite{minderer2022simple,zhou2022maskclip}. While existing works typically assume image-text pretrained models are given and focus on finetuning or adaptation strategies, our focus is to improve the upstream image-text pretraining with vision transformers.
\section{Method}
\label{sec:method}

\begin{figure*}[t]
\centering
\includegraphics[width=0.98\linewidth]{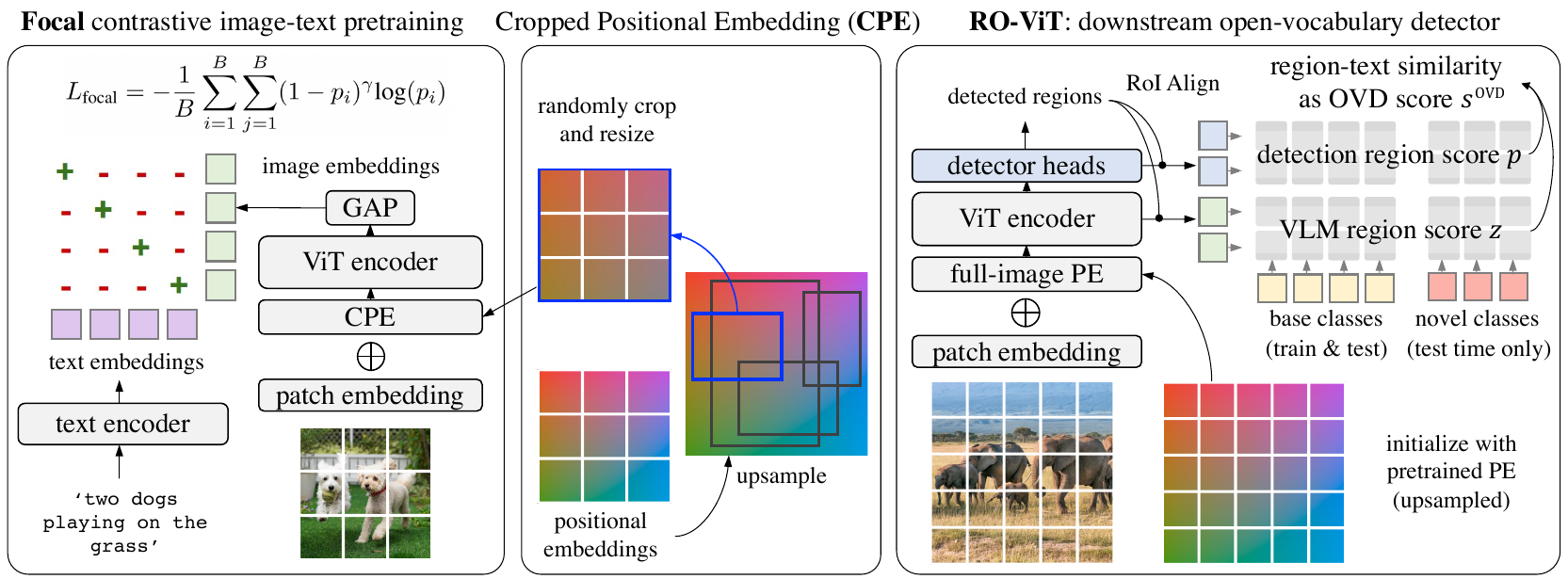}
\vspace{-1mm}
\caption{\textbf{RO-ViT framework overview.} Our Region-aware Open-vocabulary Vision Transformer (\textbf{RO-ViT}) attempts to bridge the gap between the image-level vision-language model (VLM) pretraining and downstream open-vocabulary detection. For the pretraining, we propose Cropped Positional Embedding (\textbf{CPE}) which randomly crops and resizes a \textit{region} of positional embeddings instead of using the whole-image PE. In addition, we use \textbf{focal} loss instead of the common softmax cross entropy loss for contrastive learning. The pretrained ViT backbone is transferred to the downstream open-vocabulary detection by replacing the global average pooling with detector heads. \ours detector takes the whole-image positional embeddings as input, and the detections are used to crop out region embeddings from the ViT backbone features. The region embeddings then match with the cached category embeddings to obtain the VLM score $z$, which is combined with the detection score $p$ into the open-vocabulary detection score $s^{\text{OVD}}$ (see Equation ~\ref{eqn:combine-score}).
}
\vspace{-3mm}
\label{fig:overview}
\end{figure*}

In this paper we address the problem of open-vocabulary detection. At training time, the model has access to the detection labels of $C_B$ base categories, but needs to detect objects from a set of $C_N$ novel categories at test time. We follow existing works~\cite{gu2022openvocabulary,zhong2021regionclip,du2022learning,kuo2022f} to leverage pretrained vision and language models (VLM)~\cite{radford2021clip}. Unlike existing works, we explore how to best pretrain our own VLMs with vision transformers~\cite{dosovitskiy2020image} for open-vocabulary detection.

\subsection{Preliminaries}
\label{sec:baseline}

\paragraph{Contrastive image-text pretraining.}\quad
Two-tower contrastive image-text learning involves an image encoder and a text encoder \cite{radford2021clip,align}. The text tower is typically transformer encoder, whereas the image tower can be CNN-based or ViT as in our case. Given a set of image-text corpus, the model learns to bring each image and its corresponding text together and push other non-matching texts away in the embedding space. The most common objective is softmax cross-entropy loss. The image/text embeddings are typically obtained by taking the pre-pended class token embedding, self-attention pooling, or  average pooling. We use the global average pooling followed by L2 normalization for both image and text embeddings.

\paragraph{Open-vocabulary object detector.}\quad
An open-vocabulary object detector is trained with the labels of $C_B$ base categories, but needs to detect the union of base and novel categories ($C_B \cup C_N$) at test time. Most object detectors use K-way classifiers because the number of categories are the same between train and test time. To deal with additional categories at test time, the common practice is to replace the conventional fixed-size classifier fully-connected layer with the text embeddings of base categories~\cite{Zareian_2021_CVPR,gu2022openvocabulary}. It is important that the text embeddings come from the matching text encoder of the image encoder during pretraining, so that the open-vocabulary knowledge would be preserved. We represent the background category by a ``background'' phrase and the proposals not matched to any annotations in $C_B$ are labeled as background. 

During training, we compute the detection scores $p_i$ for each region $i$ as the cosine similarity between the RoI-Align feature (\ie, region embedding) and the text embeddings of $C_B$, followed by a softmax. At test time, we expand the text embeddings from $C_B$ to $C_B \cup C_N$ plus the ``background'' category for open-vocabulary detection. We extract the VLM embedding of region $i$ by RoI-Align on the output feature map of the ViT backbone, and compute the VLM region scores $z_i$ as the cosine similarity with the $C_B \cup C_N$ text embeddings. Similarly, the detection scores $p_i$ are now computed with the $C_B \cup C_N$ text embeddings. The combined open-vocabulary detection score ${s_i}^{\text{OVD}}$ is obtained by geometric means~\cite{gu2022openvocabulary,kuo2022f}:
\begin{equation}\label{eqn:combine-score}
{s_i}^{\text{OVD}} = \begin{cases}
    z_i^{(1-\alpha)} \cdot p_i^\alpha & \text{if } i \in C_B\\
    z_i^{(1-\beta)} \cdot p_i^\beta & \text{if } i \in C_N
\end{cases}
\end{equation}
, where $\alpha, \beta \in [0, 1]$ control the weights for base and novel categories. The background score comes directly from the detection score $p_i$, because the VLM score with ``background'' phrase tends to be not as reliable.

We adopt Mask R-CNN heads in our detector and use \textit{class-agnostic} box regression and mask prediction heads following existing works~\cite{du2022learning,gu2022openvocabulary,Zareian_2021_CVPR,zhong2021regionclip,kuo2022f}. Note that we use a contrastively pretrained ViT to initialize our detector backbone, and adopt the simple feature pyramid and windowed attention to handle higher-resolution image inputs as proposed by Li~\etal~\cite{li2022exploring}.

\subsection{Region-Aware Image-Text Pretraining}
\label{sec:pretraining}
Existing vision-language models are trained to match an image as a whole to a text description~\cite{radford2021clip,align}. However, the pretraining is unaware of the alignment between its region-level representations and text tokens, which is essential to downstream open-vocabulary detection. We propose a novel Cropped Positional Embeddings (CPE) to bridge this gap, and also find it beneficial to learn from hard examples with a focal loss. Our improvements introduce no extra parameters and minimal computation costs.

\paragraph{Cropped Positional Embedding (CPE).}\quad
The positional embeddings are important to transformers as they provide the information of where each element in the set comes from. This information is often useful for downstream recognition and localization tasks.

There is a mismatch between the way the positional embeddings are used in existing contrastive pretraining approaches and open-vocabulary detection finetuning. The pretraining approaches typically apply full-image positional embeddings~\cite{radford2021clip,align} during training, and use the same positional embeddings for downstream tasks, \eg, zero-shot recognition. However, the recognition occurs at region-level for open-vocabulary detection finetuning, which requires the full-image positional embeddings to generalize to regions that they never see during the pretraining.

To bridge this gap, we propose Cropped Positional Embedding (CPE) (see center of Figure~\ref{fig:overview}). First, we up-sample the positional embeddings from the image size typical for pretraining, \eg, 224 to that typical for detection tasks, \eg, 1024. Then we randomly crop and resize a region from the up-sampled positional embeddings and use that as the image-level positional embeddings during pretraining. The regions are uniformly sampled from the normalized coordinates as $x_1 \thicksim \text{Uniform} (0, 1)$, $y_1 \thicksim \text{Uniform} (0, 1)$, $x_2 \thicksim \text{Uniform}(x_1, 1)$, $y_2 \thicksim \text{Uniform}(y_1, 1)$, while keeping the crop scale ratio in [0.1, 1.0].

Intuitively, this causes the model to view an image not as a full image in itself, but as a region crop from some larger unknown image. This better matches the downstream use case of detection where recognition occurs at region- rather than image-level.

\paragraph{Focal Loss.}\quad
We desire to have finer control over how hard examples are weighted than what the softmax cross entropy loss can provide. Focal loss offers a natural option to tune the weights of hard examples~\cite{lin2017focal} . Let $v_i$ and $l_i$ be the normalized image and text embeddings, and the image-to-text (I2T) contrastive losses be $L_\text{softmax}$ and $L_\text{focal}$ for the softmax (baseline) or focal loss (\ours). We define the losses mathematically below:
\begin{equation}
L_{\text{softmax}} = -{1 \over {B}} \sum_{i=1}^{B} \log({\text{exp}(v_{i}l_{i} / \tau) \over { \sum_{j=1}^{B} \text{exp}(v_{i} l_{j} / \tau)  }})
\label{eq:eqn1}
\end{equation}

\begin{equation}
L_{\text{focal}} = -{1 \over {B}} \sum_{i=1}^{B} \sum_{j=1}^{B} (1- p_{i})^{\gamma} \text{log}(p_{i})
\label{eq:eqn2}
\end{equation}

, where $p_i$ denotes the true class probability as below:
\begin{equation}
p_{i} =
    \begin{cases}
        \sigma(v_{i} l_{j} / \tau) & \text{if } i = j\\
        1 - \sigma(v_{i} l_{j} / \tau) & \text{if } i \neq j
    \end{cases}
\end{equation}
Here $\sigma$ denotes the sigmoid function. We adopt the simpler non alpha-balanced form of focal loss~\cite{lin2017focal}. The text-to-image (T2I) contrastive losses are symmetrical with the I2T losses by simply exchanging the inner/outer summation loops. The total loss is the sum of both I2T and T2I losses.

\subsection{Open-vocabulary Detector Finetuning}
\label{sec:detection}

Here we present two simple techniques to improve the downstream open-vocabulary detector (\secref{sec:baseline}). Despite the backbone features pretrained from the vast open-vocabulary data, the added detector layers (neck and heads) are newly trained with the downstream detection dataset (\eg, LVIS base categories). Existing approaches often miss novel/unlabeled objects in the object proposal stage because the proposals tend to classify them as background. To remedy this, we leverage recent advances in novel object proposal method~\cite{kim2022learning} and adopt the localization quality-based objectness (\ie, centerness score) instead of object-or-not binary classification score. Following~\cite{kim2022learning}, we use a single anchor per location and combine the predicted objectness score $o_{i}$ with the ensemble detection score in Equation~\ref{eqn:combine-score} to obtain the final OVD score as: ${S_i}^{\text{OVD}} = {o_{i}}^{\delta} \cdot {s_i}^{\text{OVD}.}$.

\begin{table}[t]
\centering
\small
\tablestyle{6pt}{1.12}
\begin{tabular}{l|l|l|l|l}
\multirow{2}{*}{method} & pretrained& detector & \multirow{2}{*}{\bf{AP$_r$}} & \multirow{2}{*}{\gray{AP}} \\
                      & model     & backbone  & & \\
\shline
\textbf{ConvNet based:} & & & & \\
DetPro-Cascade~\cite{du2022learning}      & ViT-B/32  & R-50          & 20.0      & \gray{27.0} \\
Detic-CN2~\cite{zhou2022detecting}           & ViT-B/32  & R-50          & 24.6      & \gray{32.4} \\
RegionCLIP~\cite{zhong2021regionclip}          & R-50x4    & R-50x4        & 22.0      & \gray{32.3} \\
ViLD-Ens~\cite{gu2022openvocabulary}            & ViT-B/32  & R-152         & 18.7      & \gray{26.0} \\
ViLD-Ens~\cite{gu2022openvocabulary}            & ViT-L/14  & EffNet-B7     & 21.7      & \gray{29.6} \\
ViLD-Ens~\cite{gu2022openvocabulary}            & EffNet-B7 & EffNet-B7     & 26.3      & \gray{29.3} \\
VL-PLM~\cite{zhao2022exploiting}              & ViT-B/32  & R-50          & 17.2      & \gray{27.0} \\
OV-DETR~\cite{zang2022open}            & ViT-B/32 & R-50     & 17.4      & \gray{26.6} \\
Rasheed~\etal~\cite{rasheed2022bridging}     & ViT-B/32  & R-50          & 21.1      & \gray{25.9} \\
PromptDet~\cite{feng2022promptdet}            & ViT-B/32 & R-50     & 21.4      & \gray{25.3} \\

\hline
\textbf{ViT based:} & & & & \\
OWL-ViT~\cite{minderer2022simple}          & ViT-H/14  & ViT-H/14      & \gray{23.3}$^*$ & \gray{35.3$^*$} \\
OWL-ViT~\cite{minderer2022simple}           & ViT-L/14  & ViT-L/14      & \gray{25.6}$^*$     & \gray{34.7$^*$} \\
\bf{\ours (ours)}    & ViT-B/16  & ViT-B/16      & {28.4}$^*$ & \gray{31.9$^*$} \\
\bf{\ours (ours)}    & ViT-L/16  & ViT-L/16      & \bf{33.6}$^*$ & \gray{36.2$^*$} \\
\bf{\ours (ours)}    & ViT-H/16  & ViT-H/16      & \bf{35.1}$^*$ & \gray{37.4$^*$} \\
\arrayrulecolor{lightgray}\hline\arrayrulecolor{black}

\bf{\ours (ours)}    & ViT-B/16  & ViT-B/16      & {28.0} & \gray{30.2} \\
\bf{\ours (ours)}    & ViT-L/14  & ViT-L/14      & {31.4} & \gray{34.0} \\
\bf{\ours (ours)}    & ViT-L/16  & ViT-L/16      & {32.1} & \gray{34.0} \\
\bf{\ours (ours)} $\dagger$ & ViT-L/16  & ViT-L/16      & \bf{32.4} & \gray{32.9} \\
\bf{\ours (ours)} & ViT-H/16  & ViT-H/16      & \bf{34.1} & \gray{35.1} \\
\end{tabular}
\vspace{-1mm}
\caption{\textbf{LVIS open-vocabulary object detection (mask AP).} \ours outperforms the best existing approach by \textbf{+7.8 AP$_r$} on novel categories. When using ViT-Base, Large and Huge \ours outperforms OWL-ViT based on ViT-Large by \textbf{+2.8}, \textbf{+8.0} and \textbf{+9.5 AP$_r$}, respectively.  All methods use the same instance-level supervision from LVIS base categories for detection training. $*$: reports {box} AP. $\dagger$: trained on LAION-2B instead of ALIGN.}
\label{tab:ovd}
\vspace{-2mm}
\end{table}

Additionally, we replace the standard classifier and mask output layer with the normalized layers~\cite{wang2021seesaw}. This L2-normalizes the weights $w$ and features $x$ as: $f(x;w,b,\tau) = {\tau \over \|w\|_2 \|x\|_2}w^{T} x + b$, where $\tau$ = 20. Although we do \textit{not} have rare categories at training (\ie, open-vocabulary setting), we empirically found it beneficial (see \secref{sec:results}).

\section{Experimental Results}
\label{sec:results}

\begin{table}[t]
\centering
\small
\tablestyle{5.0pt}{1.12}
\begin{tabular}{l|l|l|c|c}
\multirow{2}{*}{method} & pretrained & detector & \multirow{2}{*}{\bf{novel AP}} & \multirow{2}{*}{\gray{AP}} \\
                      & model     & backbone  & & \\
\shline
\textbf{ConvNet based:} & & & & \\
ViLD~\cite{gu2022openvocabulary}            & ViT-B/32  & R-50   & 27.6  & \gray{51.3} \\
OV-DETR~\cite{zang2022open}            & ViT-B/32  & R-50   & 29.4  & \gray{52.7} \\
\textbf{w/ pseudo box labels:} & & & & \\
XPM~\etal~\cite{huynh2022open}        & R-50  & R-50   & 27.0   & \gray{41.2} \\
RegionCLIP~\cite{zhong2021regionclip} $\dagger$  & R-50x4    & R-50x4     & 39.3  & \gray{55.7} \\
PromptDet~\cite{feng2022promptdet}          & ViT-B/32  & R-50   & 26.6  & \gray{50.6} \\
VL-PLM~\cite{zhao2022exploiting}            & ViT-B/32  & R-50   & 34.4  & \gray{53.5} \\
Rasheed~\etal~\cite{rasheed2022bridging} $\ddagger$   & ViT-B/32  & R-50   & 36.9  & \gray{51.5} \\
\textbf{w/ weak supervision:} & & & & \\
Detic-CN2~\cite{zhou2022detecting}          & ViT-B/32  & R-50          & 24.6      & \gray{32.4} \\
\hline
\textbf{ViT based:*} & & & & \\
\bf{\ours (ours)}    & ViT-B/16  & ViT-B/16      & {30.2} & \gray{41.5} \\
\bf{\ours (ours)}    & ViT-L/16  & ViT-L/16      &  {33.0} & \gray{47.7} \\
\end{tabular}
\vspace{-1mm}
\caption{{\textbf{COCO open-vocabulary object detection (box AP50).}} \ours represents the first ViT-based approach and demonstrates a very competitive novel AP without using pseudo labeling or weak supervision. $\dagger$: RegionCLIP uses an off-the-shelf RPN during its pretraining. $\ddagger$: Rasheed~\etal uses an external MViT detector~\cite{maaz2022class} during pretraining. *: The other ViT-based method~\cite{minderer2022simple} report their results on LVIS only.
}
\vspace{-4mm}
\label{tab:ovd_coco}
\end{table}

\paragraph{Pretraining details.}\quad
Our pretraining is performed from scratch. We adopt the ViT-B/16 and ViT-L/16 as the image encoder. The input image size is 224$\times$224 which results in 14$\times$14 positional embeddings with patch size 16$\times$16. To generate the Cropped Positional Embedding (CPE), we first interpolate the positional embeddings to size 64$\times$64. We randomly crop a region with the scale ratio in [0.1, 1.0], and the aspect ratio in [0.5, 2.0]. The region crop is resized back to the size 14$\times$14 (\ie, CPE), and is added to the patch embeddings. We use the global average pooling at the last ViT layer to obtain the image embedding. The text encoder is a 12-layer Transformer following~\cite{radford2021clip, yu2022coca}, with maximum text length 64. Both image and text embeddings are L2 normalized. While most experiments train on ALIGN~\cite{align} dataset, we report that training on the publicly available LAION-2B~\cite{schuhmann2021laion} dataset results in comparable performance (\tabref{tab:ovd}). We use a batch size 4096 for ablation and 16384 to compare with other methods. 
We use the AdamW optimizer with learning rate 5e-4 and a linear warmup of 10k steps, and train for 500k iterations.

\paragraph{Downstream detection details.}\quad
We train \ours with base categories $C_B$ for 46.1k/11.3k iterations with image size 1024, large scale jittering~\cite{ghiasi2021simple}, batch size 256/128, the SGD optimizer with weight decay 1e-4/1e-2, momentum 0.9 and an initial learning rate of 0.36/0.02 for LVIS/COCO datasets. The pretrained positional embeddings are bilinearly interpolated to adjust to the size of patch embeddings of higher resolution~\cite{dosovitskiy2020image}. We set the backbone learning rate lower (\eg, 0.1$\times$) than the rest of the model to retain the pretraining knowledge during detection finetuning. We use the score combination of ($\alpha, \beta, \delta$) = (0.65, 0.3, 3) in \secref{sec:detection}. We use CLIP~\cite{radford2021clip} prompt templates and take the average text embeddings of each category.
We use OLN-RPN~\cite{kim2022learning} at the RPN stage which uses centerness as objectness, a single anchor per location and IoU loss. The RPN NMS threshold is set to 0.7 at training and 1.0 at testing.

\begin{table*}[t]
\centering
\small
\tablestyle{6pt}{1.1}
\begin{tabular}{l|c|cccccc|cccccc}
&  image & 
\multicolumn{6}{c|}{MS COCO (5K test set)} & \multicolumn{6}{c}{Flickr30K (1K test set)} \\
& backbone & \multicolumn{3}{c}{\underline{{\white{-------}}image-to-text{\white{-------}}}} & \multicolumn{3}{c|}{\underline{{\white{-------}}text-to-image{\white{-------}}}} & \multicolumn{3}{c}{\underline{{\white{-------}}image-to-text{\white{-------}}}} & \multicolumn{3}{c}{\underline{{\white{-------}}text-to-image{\white{-------}}}} \\
method & size      & R@1  & R@5  & R@10 & R@1  & R@5   & R@10      & R@1  & R@5  & R10 & R@1  & R@5 & R@10    \\
\shline
CLIP~\cite{radford2021clip}   & 302M
& 58.4 & 81.5 & 88.1 & 37.8 & 62.4 & 72.2       & 88.0 & 98.7 & 99.4 & 68.7 & 90.6 & 95.2        \\
ALIGN~\cite{align}   & 408M
& 58.6 & 83.0 & 89.7 & 45.6 & 69.8 & 78.6       & 88.6 & 98.7 & 99.7 & 75.7 & 93.8 & 96.8        \\
FLAVA~\cite{singh2022flava}   & 86M
& 42.7 & 76.8 & - & 38.4 & 67.5 & -             & 67.7 & 94.0 & - & 65.2 & 89.4 & - \\
FILIP~\cite{yao2021filip}   & 302M
& 61.3 & 84.3 & 90.4 & 45.9 & 70.6 & 79.3       & 89.8 & 99.2 & {99.8} & 75.0 & 93.4 & 96.3        \\
Florence~\cite{yuan2021florence} & 637M
& 64.7 & 85.9 & - & 47.2&  71.4&  -             & 90.9 & 99.1 & - & 76.7 & 93.6 & -             \\
CoCa-Large~\cite{yu2022coca} & 303M
& 65.4 & 85.6 & 91.4&  50.1 & 73.8 & 81.8          & 91.4 & 99.2 & \bf{99.9} & 79.0& 95.1 & 97.4   \\
CoCa~\cite{yu2022coca} & 1B
& 66.3 & 86.2 & 91.8 & 51.2 & 74.2 & 82.0          & \bf{92.5} & \bf{99.5} & \bf{99.9} & 80.4 & 95.7 & \bf{97.7} \\
\hline
\bf{\ours (ours)} & 303M
        & \bf{68.9} & \bf{87.8} & \bf{92.2} & \bf{51.8} & \bf{75.0} & \bf{83.0}     & {92.1} & {99.4} & 99.7 & \bf{80.7} & \bf{96.1} & \bf{97.7}       \\  
\end{tabular}
\caption{\textbf{Zero-shot image-text retrieval results on COCO and Flickr30K benchmarks.} We compare with dual-encoder methods. We achieve state-of-the-art results on COCO benchmark. We outperform CoCa-Large with the same backbone by +3.5 / +1.7 R@1, and even surpass CoCa with 3$\times$ larger backbone (ViT-Giant) by +2.6 / +0.6 R@1 on image-to-text / text-to-image retrieval. We also match or outperform the state-of-the-art methods on Flickr benchmark
}
\label{tab:retrieval_sota}
\vspace{-2mm}
\end{table*}

\subsection{Open-vocabulary Object Detection}
\label{sec:ovd}

\paragraph{LVIS benchmark.}\quad We conduct evaluation on the LVIS dataset~\cite{lvis} which contains a large and diverse set of 1203 object categories suitable for open-vocabulary detection. Following the existing works~\cite{gu2022openvocabulary,zhong2021learning,du2022learning}, we use the frequent and common categories as base categories $C_B$ for training, and hold out the rare categories as novel categories $C_N$ for testing. Mask AP$_r$ is the main benchmark metric. We report the mean over three runs following~\cite{gu2022openvocabulary}.

As shown in \tabref{tab:ovd}, our best model achieves 34.1 AP$_r$, which significantly outperforms best existing ViT-based approach OWL-ViT~\cite{minderer2022simple} by \textbf{+9.5} points. Notably, our method with smaller ViT-B/16 backbone surpasses OWL-ViT with ViT-L/14 by \textbf{+2.8} AP$_r$. Compared to the best existing approach (ViLD-Ens with EffNet-B7 backbone), we outperform by +7.8 AP$_r$. We note that \ours is simply trained end-to-end with the cross-entropy loss without the employment of LVIS-tailored losses such as federated loss~\cite{minderer2022simple,zhong2021regionclip,zhou2022detecting}, weak supervision~\cite{zhou2022detecting}, or self-training~\cite{zhong2021regionclip,zhao2022exploiting,rasheed2022bridging}.

\paragraph{COCO benchmark.}\quad
We also present the comparison on the COCO benchmark where the setup uses 48 base categories for training and 17 novel categories for testing~\cite{gu2022openvocabulary}. The main metric is AP50 of novel categories. 
Due to the smaller number of training categories, we observe a tendency to overfit to these categories. Unlike most competing methods, \ours does \textit{not} use any auxiliary objectives to mitigate overfitting such as pseudo box/mask labels~\cite{huynh2022open,feng2022promptdet,zhong2021regionclip,zhao2022exploiting,rasheed2022bridging}, knowledge distillation~\cite{gu2022openvocabulary,du2022learning}, weak supervision~\cite{zhou2022detecting}. Still, \tabref{tab:ovd_coco} shows that \ours is very competitive among the other methods trained with various sources. In addition, \ours represents the first ViT-based method to be reported on this benchmark, as the other method OWL-ViT~\cite{minderer2022simple} only benchmarks on LVIS.

\subsection{Image-Text Retrieval}\quad
\label{sec:retrieval}
Apart from evaluating region-level representation through open-vocabulary detection, we evaluate the \textit{image-level} representation of \ours in image-text retrieval through the MS-COCO~\cite{chen2015cococaptions} and Flickr30K~\cite{plummer2015flickr30k} benchmarks. We further train \ours (ViT-L/16), the same model as the last row of Table~\ref{tab:ovd} and ~\ref{tab:transfer}, for 40K iterations more at a higher resolution \eg 448, following the standard of existing works~\cite{align,yu2022coca}. 

\tabref{tab:retrieval_sota} shows the comparison with other dual-encoder methods on zero-shot image-to-text (I2T) and text-to-image (T2I) retrieval. Surprisingly, \ours outperform all published works on the MS COCO benchmark, and is on par with the state of the art~\cite{yu2022coca} on Flickr. Compared to CoCa~\cite{yu2022coca} with the same backbone capacity (ViT-L), \ours outperforms on 11 out of 12 image-text retrieval metrics. Our model has 303M parameters and achieves the best performance overall. This shows that our pretraining method not only improves the region-level representation for open-vocabulary detection but also the global image-level representation for retrieval.

\begin{table}[t]
\centering
\small
\tablestyle{8pt}{1.1}
\begin{tabular}{l|l|ccc}
method              & backbone  & AP    & AP\textsubscript{50}  & AP\textsubscript{75}    \\
\shline
supervised~\cite{gu2022openvocabulary}          & R-50      & 25.6      & 38.6      & 28.0  \\
\hline
ViLD~\cite{gu2022openvocabulary}                & R-50      & 11.8      & 18.2      & 12.6  \\
DetPro~\cite{du2022learning}              & R-50      & 12.1      & 18.8      & 12.9  \\
\hline
\bf{\ours (ours)}   & ViT-B/16  & {14.0}    & {22.3}    & {14.9} \\
\bf{\ours (ours)}   & ViT-L/16  & \bf{17.1}    & \bf{26.9}    & \bf{18.5} \\
\end{tabular}
\caption{\textbf{Transfer detection on Objects365 (Box APs).} All models are trained on the LVIS base categories and tested on Objects365 dataset, without finetuning.}
\label{tab:transfer}
\vspace{-1mm}
\end{table}

\begin{table*}[t]
\centering
\subfloat[\footnotesize{\textbf{Pretraining strategy}: `base' has the standard learnable positional embeddings (PE). No PE, randomly crop and resize the last feature map (feat Crop-Resize), and sinusoidal PE are either slightly worse or help marginally. Our proposed Cropped PE (CPE) improves by +2.4 AP$_r$. Focal contrastive loss brings the gain further to +2.9 AP$_r$.\label{tab:ablation:pretraining}}]{
\tablestyle{6.0pt}{1.1}\begin{tabular}{l|cc|c|c}
pretraining method &   \bf{CPE} & \bf{focal} & AP$_r$  &   \gray{AP} \\
\shline
base             &&& 21.4 {\white{(+0.0)}}         & \gray{26.6}  \\
no PE            &&& 18.7 {\white{(+0.0)}} & \gray{25.2}  \\
SinCos PE        &&& 21.7 {\white{(+0.0)}}  &  \gray{26.9}\\
feat Crop-Resize &&& 20.6 {\white{(+0.0)}}  &  \gray{26.6} \\
\hline
ours         &\checkmark&& 23.8 ({\green{+2.4}})  & \gray{27.4}  \\
ours         &&\checkmark& 21.9 ({\green{+0.5}})  & \gray{27.4}  \\
ours         &\checkmark&\checkmark& 24.3 ({\green{+2.9}})  & \gray{27.6}  \\
\end{tabular}}\hspace{4mm}
\subfloat[\footnotesize{\textbf{Frozen backbone study}: Only the newly added detector layers are trained in detection finetuning, to directly evaluate the pretrained features for open-vocabulary detection. Our proposed CPE improves the baseline by a large margin of +6.5 AP$_r$.\label{tab:ablation:frozen}}]{
\tablestyle{7.0pt}{1.1}\begin{tabular}{l|c|c}
frozen backbone &   AP$_r$  &   \gray{AP} \\
\shline
base         & 9.7 {\white{(+0.0)}}   & \gray{12.2}  \\
\hline
CPE          & 16.2 ({\green{+6.5}})  & \gray{17.1}  \\
CPE + focal  & 16.5 ({\green{+6.8}})  & \gray{17.1}  \\
 \multicolumn{3}{c}{~}\\
 \multicolumn{3}{c}{~}\\
 \multicolumn{3}{c}{~}\\
 \multicolumn{3}{c}{~}\\
\end{tabular}}\hspace{4mm}
\subfloat[\footnotesize{Ablating \textbf{backbone finetuning learning rate ratio} (`bblr') w.r.t. added detector layers.\label{tab:ablation:bblr}}]{
\tablestyle{6.0pt}{1.1}\begin{tabular}{l|c|c}
bblr &   AP$_r$  &   \gray{AP} \\
\shline
0.0         & 16.5   & \gray{17.1}  \\
0.001                   & 19.7   & \gray{22.9}  \\
0.01                   & 20.4   & \gray{23.0}  \\
0.1          & 24.3   & \gray{27.6}  \\
1.0          & 17.9   & \gray{25.1}  \\
 \multicolumn{3}{c}{~}\\
 \multicolumn{3}{c}{~}\\
\end{tabular}}\vspace{2mm}
\subfloat[\footnotesize{\textbf{Downstream detector improvement:} Combining the localization quality-based objectness from the proposal stage to the detection scoring (`loc.obj.') improves by $\sim$2 AP$_r$. Normalized classifier and mask output layer (`norm.lyr') gives additional gain.}\label{tab:ablation:detector}]{
\tablestyle{8.0pt}{1.1}\begin{tabular}{l|cc|c|c}
backbone    & loc.obj.  & norm.lyr   &    AP$_r$  &   \gray{AP} \\
\shline
ViT-B/16    &           &           & 21.4 {\white{(+0.0)}}         & \gray{26.6}  \\
ViT-B/16    &\checkmark &           & 23.4 ({\green{+2.0}})   & \gray{27.2}  \\
ViT-B/16    &           &\checkmark & 22.0 ({\green{+0.6}})   & \gray{26.8}  \\
ViT-B/16    &\checkmark &\checkmark & 23.9 ({\green{+2.5}})   & \gray{27.5}  \\
\hline
ViT-L/16    &           &           & 25.1 {\white{(+0.0)}}         & \gray{30.1}  \\
ViT-L/16    &\checkmark &\checkmark & 27.2  ({\green{+1.9}})  & \gray{30.6}  \\
 \multicolumn{5}{c}{~}\\
 \multicolumn{5}{c}{~}\\
\end{tabular}}\hspace{4mm}
\subfloat[\footnotesize{\textbf{Scalability w.r.t model size and contrastive batch size:} The benefit of our `CPE + focal' holds with larger model and batch size, improving AP$_r$ up to +2.7 points. `imp.(d)' denotes the detector improvements in sub-table (d).}\label{tab:ablation:scale}]{
\tablestyle{7pt}{1.1}\begin{tabular}{cc|c|c|c|c}
backbone    &   batch  &  imp.(d) & \textbf{CPE + focal} &  AP$_r$  &   \gray{AP} \\
\shline
ViT-B/16    & 4096     &           &           & 21.4  {\white{(+0.0)}}          & \gray{26.6}  \\
\hline
ViT-B/16    & 4096     &\checkmark &           & 23.9  {\white{(+0.0)}}          & \gray{27.5}  \\
ViT-B/16    & 4096     &\checkmark &\checkmark & 26.2  ({\green{+2.3}})         & \gray{29.2}   \\
\hline
ViT-B/16    & 16384    &\checkmark &           & 26.4 {\white{(+0.0)}}          & \gray{30.3}   \\
ViT-B/16    & 16384    &\checkmark &\checkmark & 28.0  ({\green{+1.6}})         & \gray{30.2}   \\
\hline
ViT-L/16    & 4096     &\checkmark &          & 27.2  {\white{(+0.0)}}         & \gray{30.6}  \\
ViT-L/16    & 4096     &\checkmark &\checkmark & 29.9  ({\green{+2.7}})         & \gray{31.1}  \\
\hline
ViT-L/16    & 16384    &\checkmark &\checkmark & 32.1 {\white{(+0.0)}}          & \gray{34.0}  \\
\end{tabular}}
\caption{\textbf{Ablation studies.} We follow LVIS open-vocabulary detection benchmark. We train on base (`frequent' + `common') categories, test on novel (`rare') categories, and report AP$_r$. We use ViT-B/16 backbone and contrastive batch size 4096 unless otherwise noted.}
\label{tab:ablations}
\vspace{-1mm}
\end{table*}

\subsection{Transfer Object Detection}
\label{sec:transfer}
We evaluate the generalization ability of \ours in zero-shot transfer detection. We use the same detector trained on the LVIS base categories (\secref{sec:ovd}) and test on Objects365-v1 validation split~\cite{objects365} following the setup of ViLD~\cite{gu2022openvocabulary,du2022learning}. We replace the LVIS  with Objects365 vocabulary embeddings to perform the transfer detection without finetuning.

\tabref{tab:transfer} summarizes the box AP scores in comparison with prior works. Our best model achieves 17.1 AP, which outperforms existing works ViLD by +5.3 AP and DetPro by +5.0 AP. Due to the different backbone capacity (R50 vs ViT), we note that this comparison is to showcase that \ours can achieve strong cross-dataset detection generalization. For transfer detection, we assume all categories are novel and set $\alpha, \beta$=(0.0, 0.65) in Equation~\ref{eqn:combine-score}.

\subsection{Ablation Study}
\label{sec:ablation}
We provide ablation studies on the LVIS open-vocabulary detection and image-text retrieval benchmarks.

\paragraph{Pretraining strategy.}\quad
\tabref{tab:ablation:pretraining} ablates our proposed methods in the contrastive image-text pretraining. We start with ``base'' that uses the common full-image positional embeddings and cross entropy loss. We explore different PE and observe that removing the positional embeddings hurts the performance (no PE), while sinusoidal PE (SinCos PE) yields a similar performance to the baseline. We also try the random crop and resize on the last backbone feature map (``feat Crop-Resize") to use region features during pretraining but see no improvement. 
In contrast, our proposed Cropped Positional Embedding (CPE) offers a clear benefit of +2.4 AP$_r$ and focal contrastive loss (focal) yields additional gains. Our method achieves a gain of +2.9 AP$_r$ in total. This shows that CPE can learn strong region-level representations, and our focal contrastive learning helps preserving the knowledge about the pretrained image-text concepts, both of which are essential for open-vocabulary detection.

\begin{table}[t]
\centering
\tablestyle{7pt}{1.1}\begin{tabular}{lcc|cccc}
\multirow{2}{*}{backbone}    & 
\multirow{2}{*}{focal}   & 
\multirow{2}{*}{CPE}  &
\multicolumn{2}{c}{MS COCO} & \multicolumn{2}{c}{Flickr30K}\\
& & & I2T & T2I & I2T & T2I \\
\shline
ViT-B/16 &          &           & 59.1 & 42.5 & 84.8 & 70.9  \\
ViT-B/16 & \checkmark &          & 60.3 & 44.0 & 85.4 & 71.6  \\
ViT-B/16 & \checkmark & \checkmark & 62.2 & 44.2 & 86.5 & 71.8  \\
ViT-\textbf{L}/16 & \checkmark & \checkmark & 67.0 & 49.7 & 89.5 & 77.2 \\
\end{tabular}
\vspace{-1mm}
\caption{\textbf{Pretraining evaluation on zero-shot image-text retrieval (Recall@1).} We evaluate the image-level representation of our pretrained model on COCO and Flickr30k retrieval tasks. We ablate the focal contrastive loss, Cropped Positional Embedding (CPE) and backbone size.}
\label{tab:retrieval}
\vspace{-2mm}
\end{table}

\paragraph{Image-text retrieval.}\quad
In \tabref{tab:retrieval}, we demonstrate that our proposed focal contrastive loss and Cropped Positional Embedding (CPE) both improve the image-text retrieval. We use a batch size of 16384 , image size 224, and ViT-B/16 backbone as our baseline, and report Recall@1 metrics on MS COCO and Flickr datasets. We first observe that focal loss brings clear improvements on all metrics, and adding CPE brings further improvements with a total of +3.1 / +1.7 image-to-text (I2T) and +1.7 / +0.9 text-to-image (T2I) improvements on the MS COCO / Flickr datasets. By replacing ViT-B/16 with ViT-L/16, we observe marked improvements of +4.8 / +5.5 / +3.0 / +5.4 on these metrics, showing that our recipe is highly scalable with backbone capacity.

\paragraph{Frozen backbone study.}\quad
We investigate the effectiveness of the \textit{frozen} backbone features in open-vocabulary detection in \tabref{tab:ablation:frozen}. This evaluation setting is more rigorous for representation learning because finetuning an entire network evaluates not only the quality of the representations but also the initialization and optimization method, as discussed in Goyal~\etal~\cite{goyal2017accurate}. During the detection finetuning, all ViT backbone layers are frozen, and only the added detector layers (neck and heads) are trained. We observe frozen \ours backbone improves the baseline by +6.8 AP$_r$, a much larger margin compared to the full-finetuning setting. This study shows the region-aware representation of \ours is critical for downstream detection tasks.

\paragraph{Backbone learning rate ratio.}\quad
As discussed in~\secref{sec:baseline}, our open-vocabulary object detector depends on the pretrained knowledge in the backbone to recognize novel categories beyond the detector annotations. Therefore, it is important to set the backbone learning rates lower than the rest of the detector layers to prevent forgetting in the finetuning process. We present ablations on the learning rate ratio between the backbone and the detector layers in Table~\ref{tab:ablation:bblr}. We found 0.1$\times$ to be the sweet spot. Higher ratios lead to forgetting and hurts the AP$_r$, and lower ratios hurt the ability to adapt to detection tasks and hurts AP overall.

\paragraph{Downstream detector improvements.}\quad
In addition, \tabref{tab:ablation:detector} demonstrates our improvements in the downstream detector. Learning the objectness by localization quality, \ie, centerness (``loc.obj.")~\cite{kim2022learning} and leveraging it into detection score improves AP$_r$ by a clear gain of +1.9 points compared to using the conventional binary-classification in the proposal. This indicates the novel, unlabeled objects are indeed often missed in the proposal stage, which may limit the final open-vocabulary detection performance. Localization-quality based objectness helps remedy this.
The use of normalized activation in the last layers of the classifier and mask heads additionally improves the performance.

\paragraph{Scalability with respect to model size and batch size.}\quad
While increasing the model capacity and batch size have been beneficial for contrastive learning in general~\cite{basic,radford2021clip,yu2022coca,align}, we study their benefits in the downstream open-vocabulary detection in \tabref{tab:ablation:scale}. Note the detector improvements presented in \secref{sec:detection} are applied. We first observe increasing the batch size from the default 4096 to 16384 shows gains of +2.1$\sim$2.5 AP$_r$ for both ViT-B/L. Then, we notice upgrading ViT-B to ViT-L brings +3.3$\sim$3.7 AP$_r$ gain across different batch sizes. Last but not least, the gain of our proposed ``CPE + focal'' pretraining is consistent among all settings by improving +2.2$\sim$2.7 AP$_r$. Putting everything together, \ours achieves 32.1 AP$_r$.

\begin{figure}[t]
\centering
\includegraphics[width=1.0\columnwidth]{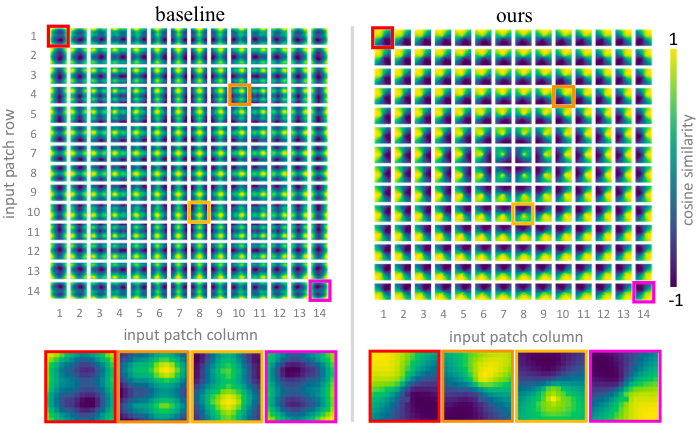}
\vspace{-5mm}
\caption{\small \textbf{Visualization of learned positional embeddings.} Tiles show the cosine similarity between the positional embedding of the patch (at the indicated row-column position) and the positional embeddings of all other patches. ViT-B/16 backbone is used.}
\label{fig:pe}
\vspace{-2mm}
\end{figure}

\subsection{Visualization of Positional Embeddings}
In \figref{fig:pe}, we visualize and compare the learned positional embeddings of \ours with the baseline, based on ViT-B/16 backbone. Each tile is the cosine similarity between positional embeddings of one patch and all other patches. The brightness patterns show in both models, closer patches have more similar positional embeddings, indicating the encoded distance and locality within the image.

We observe a few differences between \ours positional embeddings and the baselines. First, \ours forms more distinct clusters at different patch locations, \ie, ours shows symmetrical global pattern around the center patch, while the baseline has similar patterns on opposite ends of the image (\eg, the pattern in top-left patch is similar to the bottom-right patch).
Also, the brightness patterns of \ours tend to be more concentrated and strong (\ie, near 1 or -1). To summarize, the visualization indicates \ours positional embeddings acquire more structure and symmetry than the baseline in the pretraining process. 

\section{Conclusion}
\label{sec:conclusion}
We present \ours -- a contrastive image-text pretraining framework to bridge the gap between image-level pretraining and open-vocabulary detection finetuning. Our methods are simple, scalable, and easy to apply to any contrastive backbones with minimal computation overhead and no increase in parameters. Experiments show that our \ours achieves the state-of-the-art on LVIS open-vocabulary detection benchmark. Moreover, \ours achieves state of the art on 9 out of 12 metrics of image-text retrieval benchmark, showing that the learnt representation is not only beneficial at region-level but also highly effective on image-level. We hope this study can help the research on open-vocabulary detection from the perspective of image-text pretraining which can benefit both region-level and image-level tasks.

\clearpage
\newpage

{\small
\bibliographystyle{ieee_fullname}
\bibliography{egbib}
}

\clearpage
\appendix
\addcontentsline{toc}{section}{Appendices}
\section*{Appendix}
In the supplementary materials, we provide our detection visualizations along with our application on ego-centric data. We also provide more implementation details with used hyper-parameters and discuss the current limitations in the proposed \ours in the hope to inspire more future research.

\section{Implementation Details}
\tabref{tab:hparams} summarizes the hyper-parameters used in the image-text pretraining and open-vocabulary detection finetuning.

\section{Detection Visualization}
\label{sec:detection_vis}
\begin{figure*}[t]
\centering
\includegraphics[width=0.7\linewidth]{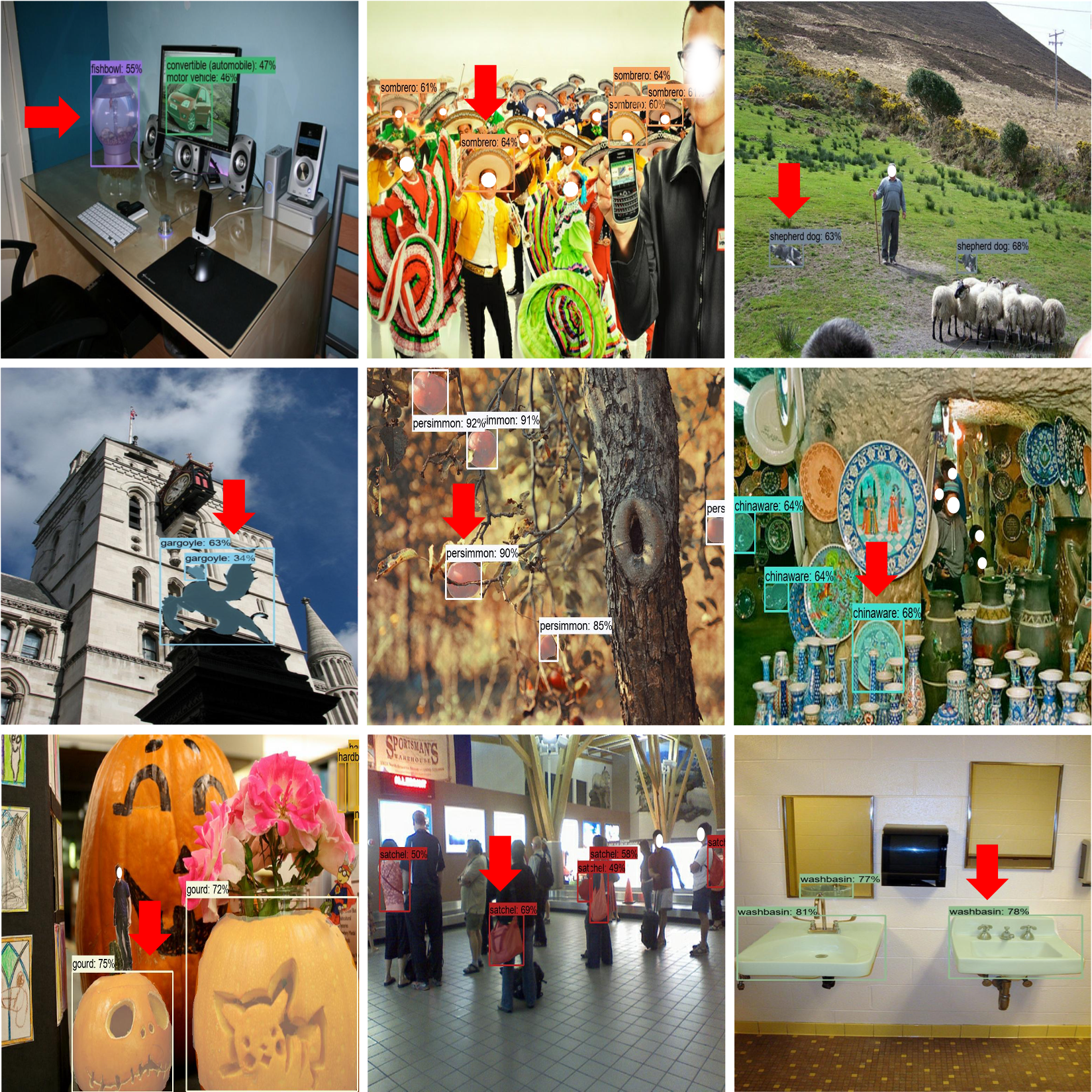}
\vspace{-2mm}
\caption{\small \textbf{LVIS novel category visualization (prediction).} We only show the novel categories for clarity. \ours detects many novel categories (pointed by the red arrows) that it has never seen during detection training (\eg, \textit{fishbowl, sombrero, shepherd dog, gargoyle, persimmon, chinaware, gourd, satchel}, and \textit{washbasin}). }
\label{fig:lvis}

\vspace{1mm}

\centering
\includegraphics[width=0.7\linewidth]{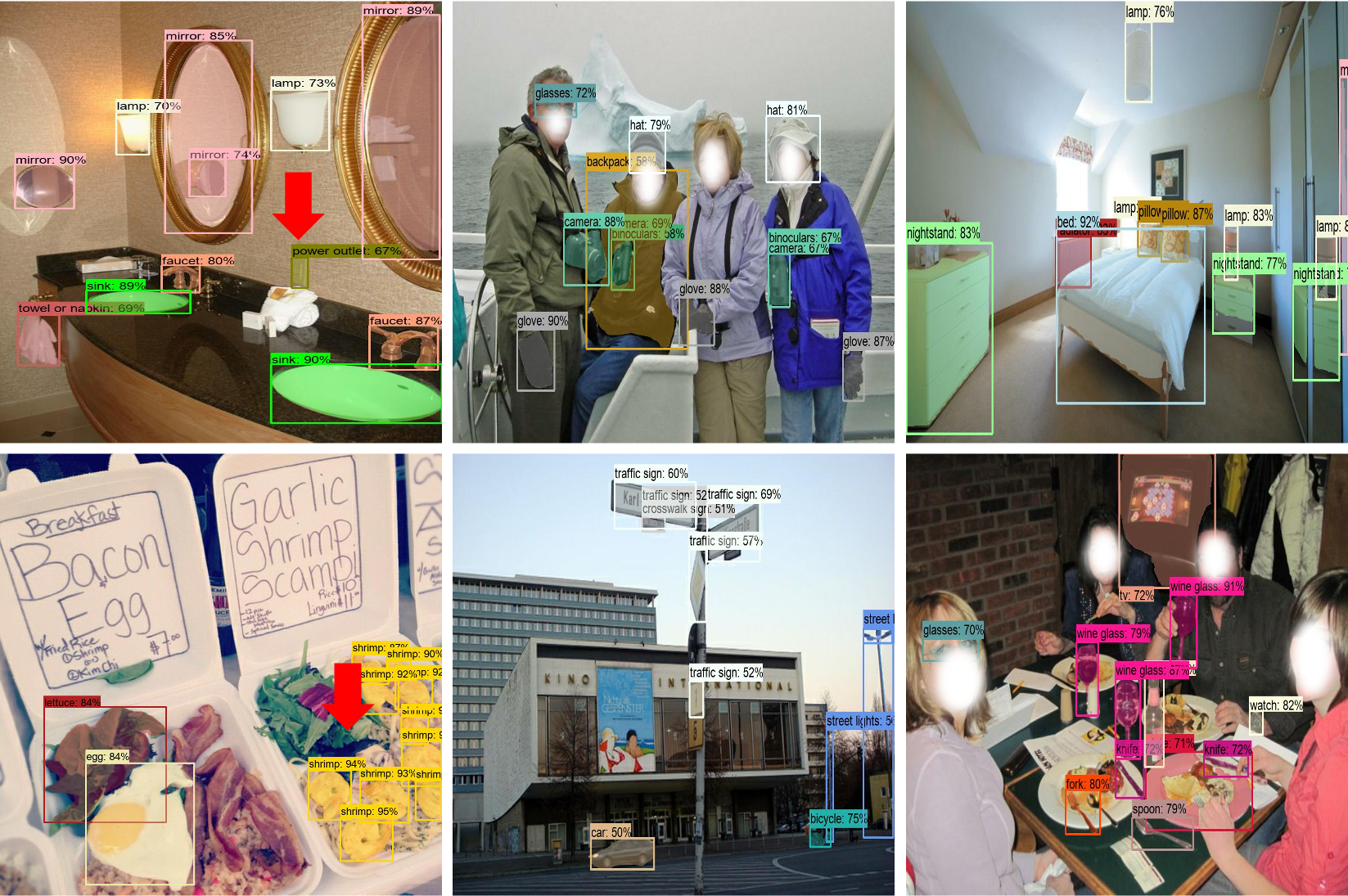}
\vspace{-2mm}
\caption{\small \textbf{Objects365 transfer detection visualization (prediction).} Our trained \ours is able to perform on a new dataset without any finetuning, and captures many challenging categories including novel categories (pointed by red arrows, \eg, \textit{power outlet} and \textit{shrimp}).}
\label{fig:obj365}
\end{figure*}

We visualize our \ours outputs on LVIS novel categories (\secref{sec:ovd}) and transfer detection (\secref{sec:transfer}) onto Objects365 in \figref{fig:lvis} and \ref{fig:obj365}, respectively. 

We use the ViT-B/16 backbone for visualization. The model was trained on the LVIS base categories following \secref{sec:ovd} of the paper. On LVIS, we only show the novel categories for clarity. \ours is able to detect many novel objects (\eg, \textit{fishbowl, sombrero, shepherd dog, gargoyle, persimmon, chinaware, gourd, satchel}, and \textit{washbasin}). We also visualize the transfer detection on Objects365 by replacing the vocabulary without finetuning. \ours can detect a wide range of objects in complex scenes (\eg, \textit{power outlet, binocular, glasses, traffic sign}, and \textit{shrimp}).

\section{Application on Ego-centric Data}
A main advantage of open-vocabulary detection is to deal with out-of-distribution data with categories given by the users on the fly. We test \ours's transfer detection to a real-world ego-centric data, Ego4D~\cite{grauman2022ego4d}. We use the same \ours trained on LVIS base categories with ViiT-B/16 backbone, as in \secref{sec:detection_vis}, \ie, the model has been never trained on Ego4D.

The categories are provided by the user's visual inspection of the video, and are as follows.

\begin{itemize}
    \item For the indoor scene: \textit{plate, cabinet, stove, towel, cleaning rag, ventilator, knob, sauce and seasoning, steel lid, window, window blinds, plant, light switch, light, door, carpet, exit sign, doormat, hair, door lock, tree, poster on the wall, sticker on the wall, faucet, recycle bin, rack, hand, can, carton, trash, Christmas tree, plastic container, fridge}.
    \item For the grocery store scene: \textit{exit sign, poster, chocolate bar, bag of candy, bag of cookies, snack, oreo, soy sauce, apple, pear, orange, grapes, price tag, cereal, instant noodle/ramen, cracker, ATM machine, instant noodle, wooden basket, red ramen bowls, magazine, drugs and medicine, Mayo, Ketchup, Cup noodle, burrito, Lays/Sun chips, seasoning sauce, black carton, salad dressing, canned food}.
\end{itemize}

\begin{figure*}[t]
\centering
\includegraphics[width=1.0\linewidth]{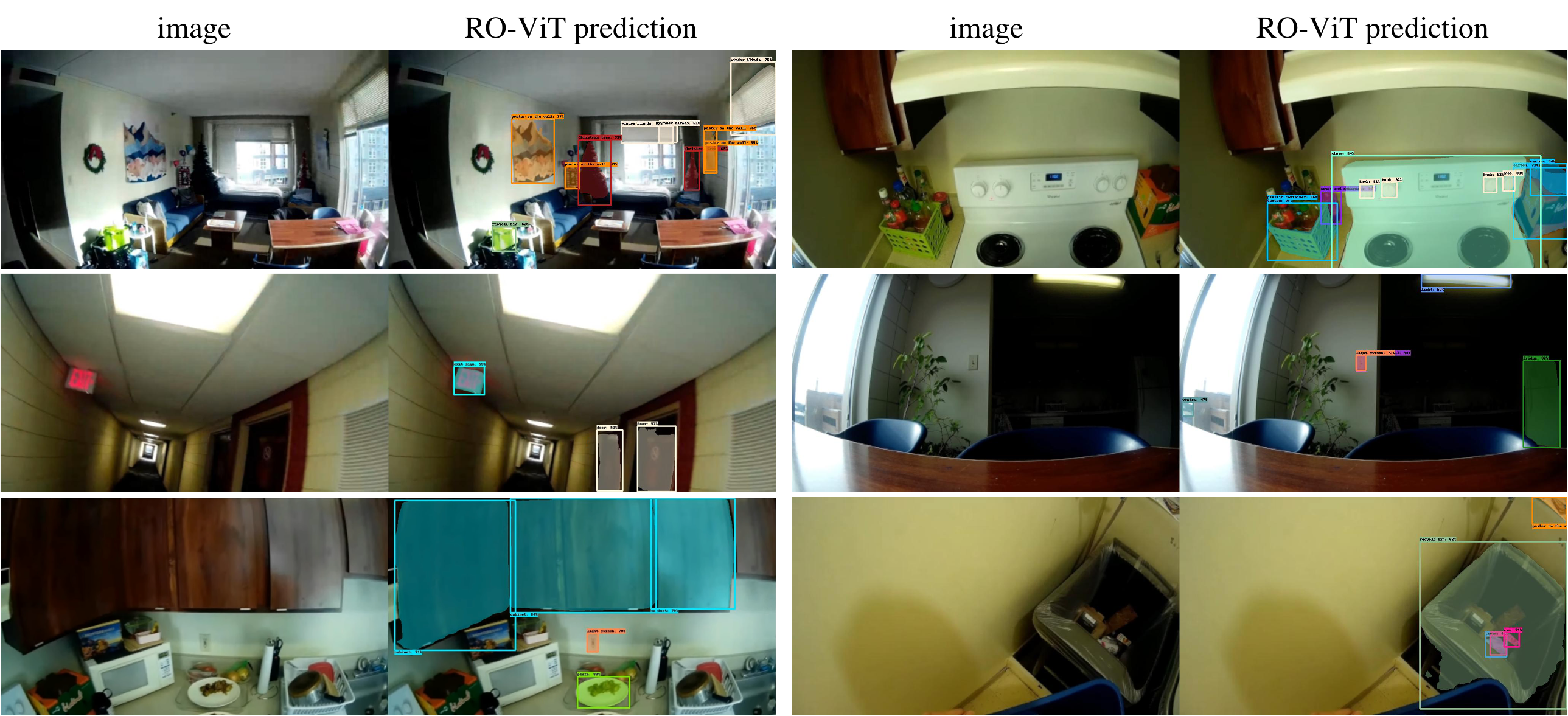} \\
(a) \bf{Indoor scene.} \\
\includegraphics[width=1.0\linewidth]{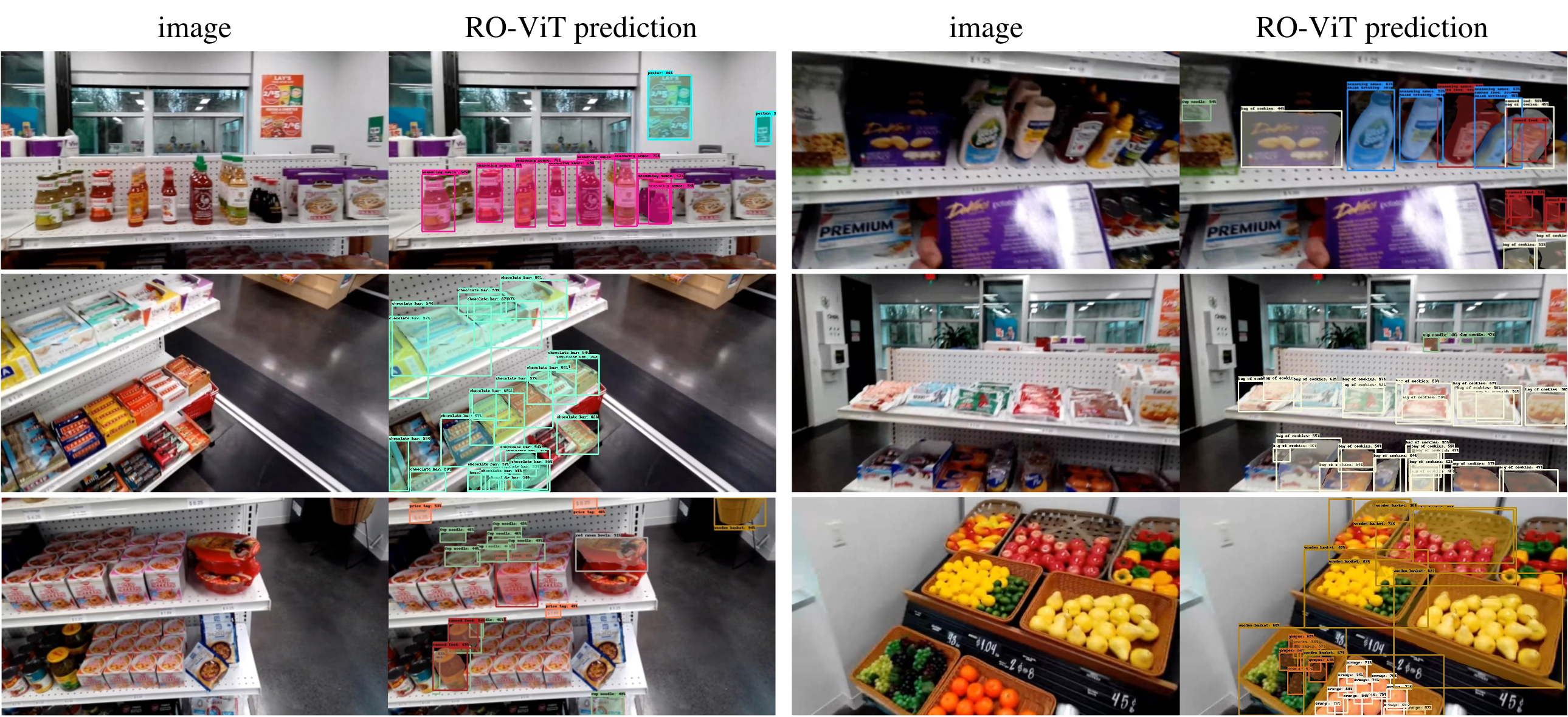} \\
(b) \bf{Grocery store scene.} \\
\caption{\small \textbf{Ego4D transfer detection visualization (prediction).} Ego4D~\cite{grauman2022ego4d} is a real-world and out-of-distribution data. Despite large domain shift and heavy camera movement, \ours is able to detect novel, unseen objects (\eg, \textit{light switch, exit sign, recycle bin, seasoning sauce, salad dressing, bag of cookies, canned food, and red ramen bowls}).}
\label{fig:ego4d}
\end{figure*}

\figref{fig:ego4d} shows our \ours prediction. Despite the large domain shift and heavy camera motions, \ours is able to capture many objects in the ego-centric videos.   Specifically, it is able to detect many novel categories never seen during training (\eg, \textit{light switch, exit sign, recycle bin, seasoning sauce, salad dressing, bag of cookies, canned food, and red ramen bowls}).

\begin{table}[t]
\centering
\small
\tablestyle{2pt}{1.1}
\begin{tabular}{l|c|c}
\multirow{2}{*}{configuration}
 & contrastive  & open-vocab. detection  \\
 & image-text pretraining  & finetuning (LVIS)   \\
\shline
optimizer           & AdamW         & SGD       \\
momentum            & $\beta$=0.9   & $\beta$=0.9       \\
weight decay        & 1e-2          & 1e-4      \\
learning rate       & 5e-4 (linear decay) & 0.36 (B) / 0.18 (L)     \\
step decay factor   & -             & 0.1$\times$ \\
step decay schedule & -             & [0.8, 0.9, 0.95]  \\
backbone lr ratio   & N/A           & 0.1 (B) / 0.5 (L) \\
warmup steps        & 1e4           & 1k        \\
total steps         & 5e5           & 46.1k (B) / 36.8k (L)\\
batch size          & 4096 or 16384 & 256 (B) / 128 (L)      \\
image size          & 224           & 1024      \\
\end{tabular}
\caption{\textbf{\ours hyper-parameters} for image-text pretraining and open-vocabulary detection finetuning. B and L denote ViT-B/16 and ViT-L/16 backbones respectively.}
\label{tab:hparams}
\end{table}

\section{Limitations}
\ours leverages the knowledge in pretrained Vision Language Models (VLM). Therefore, the biases of trained VLMs can proparate into the downstream detector. In this paper, we use \ours to demonstrate its capabilities and compare with existing works in open-vocabulary detection. We recommend careful analysis of ethical risks before using it for other purposes.
\section{Dataset License}

\begin{itemize}
    \item LVIS~\cite{lvis}: CC BY 4.0 + COCO license
    \item COCO Captions (retrieval)~\cite{chen2015cococaptions}: CC BY
    \item Flickr30k (retrieval)~\cite{plummer2015flickr30k}: Custom (research-only, non-commercial)
    \item Objects365~\cite{objects365}: Custom (research-only, non-commercial)
    \item Ego4D~\cite{grauman2022ego4d}: \url{https://ego4d-data.org/pdfs/Ego4D-Licenses-Draft.pdf}
\end{itemize}

\end{document}